\documentclass[11pt, conference]{IEEEtran}
\IEEEoverridecommandlockouts
\usepackage{hyperref}
\usepackage[a4paper,left=0.625in,right=0.625in,top=0.75in,bottom=1in]{geometry}

\usepackage{amsmath} 
\usepackage{amssymb}  
\usepackage{mathtools}
\usepackage{upgreek}
\usepackage{color,soul}
\usepackage{lipsum}
\usepackage{float}
\usepackage{balance}
\usepackage{url}
\usepackage{xr}
\usepackage{algpseudocode}
\usepackage{xcolor}
\usepackage{breqn}

\usepackage{booktabs}
\usepackage{cite}
\usepackage{graphicx}
\usepackage{graphics} 
\usepackage{subcaption} 
\usepackage{epsfig} 
\usepackage{gensymb}
\usepackage[ruled,longend,linesnumbered]{algorithm2e}
\algnewcommand{\algorithmicand}{\textbf{ and }}
\algnewcommand{\algorithmicor}{\textbf{ or }}
\algnewcommand{\algorithmicnot}{\textbf{ not }}
\algnewcommand{\OR}{\algorithmicor}
\algnewcommand{\AND}{\algorithmicand}
\algnewcommand{\NOT}{\algorithmicnot}

\usepackage{amssymb}
\usepackage{amsthm}
\usepackage{placeins}

\algdef{SE}[VARIABLES]{Variables}{EndVariables}
   {\algorithmicvariables}
   {\algorithmicend\ \algorithmicvariables}
\algnewcommand{\algorithmicvariables}{\textbf{global variables}}
\algdef{SE}[VARIABLESDEF]{VariablesDef}{EndVariablesDef}
   {\algorithmicvariablesdef}
   {\algorithmicend\ \algorithmicvariablesdef}
\algnewcommand{\algorithmicvariablesdef}{\textbf{variables definition}}

\newcommand{\density}{p}

\newcommand{\state}{\mathbf{s}}

\newcommand{\stp}{{\state_{t+1}}}

\newcommand{\pdyn}{\density}

\newcommand{\reward}{r}

\newcommand{\V}{V}

\newcommand{\Q}{Q}

\newcommand{\policy}{\pi}

\newcommand{\params}{\theta}

\newcommand{\pparams}{{\phi}}   
\newcommand{\vparams}{{\psi}}   
\newcommand{\vtargetparams}{{\bar\psi}}   

\newcommand{\discount}{\gamma}

\title{Intelligent DRL-Based Adaptive Region of Interest for Delay-sensitive
	Telemedicine Applications}
\author{\author{Abdulrahman Soliman$^{*}$, Amr Mohamed$^{*}$, Elias Yaacoub$^{*}$, 
		\\Nikhil V. Navkar${\ddag}$, Aiman Erbad$^{\dag}$\\
		$^*$Department of Computer Science and Engineering, Qatar University, Qatar \\ $^\ddag$Department of Surgery, Hamad Medical Corporation, Qatar\\
		$^\dag$College of Science and Engineering, Hamad Bin Khalifa University, Qatar  \\
		Email: abdulrahman.sulaiman@qu.edu.qa, \{amrm, eliasy, aerbad\}@ieee.org,   nnavkar@hamad.qa\\
  		
	}
}

\begin{document}

\maketitle

\pagestyle{plain}    

\begin{abstract}
	Telemedicine applications have recently received substantial potential and interest, especially after the COVID-19 pandemic. Remote experience will help people get their complex surgery done or transfer knowledge to local surgeons, without the need to travel abroad. Even with breakthrough improvements in internet speeds, the delay in video streaming is still a hurdle in telemedicine applications.
	This imposes using image compression and region of interest (ROI) techniques to reduce the data size and transmission needs. This paper proposes a Deep Reinforcement Learning (DRL) model that intelligently adapts the ROI size and non-ROI quality depending on the estimated throughput.
	The delay and structural similarity index measure (SSIM) comparison are used to assess the DRL model. The comparison findings and the practical application reveal that DRL is capable of reducing the delay by 13\% and keeping the overall quality in an acceptable range. Since the latency has been significantly reduced, these findings are a valuable enhancement to telemedicine applications.

\end{abstract}

\begin{IEEEkeywords}
	\makeatletter
	\def\RemoveSpaces#1{\zap@space#1 \@empty}
	\makeatother
	Telemedicine, Deep Reinforcement Learning (DRL), region of interest (ROI), optimization, structural similarity index measure (SSIM).
\end{IEEEkeywords}

\section{Introduction}
\label{sec:intro}
Telemedicine is defined as the provision of various healthcare services over a telecommunication network \cite{tele}. Telesurgery and surgical tele-mentoring are two examples of surgical field telemedicine. Flying overseas to well-known surgeons is usually the only option for people who need sophisticated surgical operations. By using telemedicine, healthcare practitioners will get physical or virtual experience from specialists abroad for surgical procedures, saving patients time and money. Transmitting real-time instructions or high definition (HD) video streams is a challenge due to the large average delay, which can reach to 260±44 ms \cite{Shabir}.

The ultimate objective is to reduce latency and transmit in real-time, which requires either very high bandwidth or an adaptive technique to remove unneeded information when bandwidth is limited. The applications of surgical tele-mentoring are the main focus of this work, as in \cite{Shabir,Somayya}. Tele-mentoring is the sharing of experience and procedures between overseas specialists and local operators leveraging telecommunications. To fully utilize this technology, very high bandwidth is required, resulting in very short delays and interactions that appear to be in real-time. Because stable high bandwidth is unavailable in certain countries or at all times, adopting transmission techniques and controlling quality is an inevitable procedure to compensate for the low bandwidth issue.
Optimization solutions and experiments have been done in \cite{Somayya}, and the results have shown some potential in reducing the delay and improving the quality. Still, this solution consumes time and is not practical in real-time, because it must determine the best ROI size and quality factor for the throughput input at each timestep. It also did not take into account the dynamic change of throughput, which is a key characteristic of throughput.
We extend the work in \cite{Somayya} by presenting a DRL model that adapts the ROI size and non-ROI quality based on the current and expected throughput. The model aims to get maximum rewards by selecting the best ROI size and non-ROI quality that fits the current throughput. The regression model contributes to the calculation of the delay by obtaining the frame size in bytes using the ROI dimensions and the discrete cosine transform (DCT) quality factor (QF). To evaluate the quality, SSIM was utilized to compare the difference in quality between the original and compressed frames. 

\section{Related Work}
\label{sec:related}
Several works have been published to reduce the delay or compress the frames, but due to the visible latency that impedes real-time transmission, these works are insufficient to accept and permit in practical telemedical applications.

The introduction of a low-complexity image compression approach for digital imaging and communications in medicine (DICOM) was documented in [4]. The authors of the study have effectively employed the primary benefits of the region-based coding approach in their research. The region of interest (ROI) is manually detected and combined with the effects of the Integer Wavelet Transform (IWT). This technique demonstrates utility in the reversible reconstruction of the original frame, achieving the appropriate level of quality. The overall compression process plays a significant role in attaining an adequate degree of picture transmission within limited bandwidth constraints in the context of telemedicine applications. However, a limitation of this study is the reliance on human ROI selection. Another study was conducted by researchers in [5], which aimed to decrease the size of frames. In this study, the researchers developed an image compression method for MR images that focuses on an ROI and takes into account the frequency components of the medical picture being processed. The Fuzzy C-Means clustering technique separates the ROI from the non-ROI regions. The technique of capsule autoencoder is employed for compressing the non-ROI, whilst the Discrete Cosine Transform with Huffman Run-length encoding is employed for compressing the area of interest.

Path quality and latency minimization algorithms were implemented in \cite{Chapagain}. This is accomplished by predicting data delivery time and packet drop probability in the route quality computation. The authors were able to minimize the processing time and end-to-end latency by an average of 32.6 ms and 33.93 ms, respectively. This method is effective, but it delays real-time video stream transmission due to complicated computations and data delivery time prediction. 

Work in \cite{Somayya} is the most recent and closely related work to ours, and it covered a wide range of system aspects, from identifying ROI to regression and optimization models. The authors began by discussing different ROI detection methods, such as segmentation and k-means, but they concluded that a shallow convolutional neural network (S-CNN) is the most effective method to adopt owing to its speed and efficiency.
The optimization model, which establishes a multi-objective optimization function to optimize overall quality while minimizing delay, is another segment they proposed. Despite encouraging results in terms of reduced latency and improved quality, this approach cannot be used or implemented in real-world telemedicine applications. According to  \cite{Liao}, multi-objective optimization methods require a significant amount of computation time to provide the best parameters; this delay is unacceptable in delay-sensitive applications such as telesurgery because waiting for an optimization function to produce the size of the ROI for the current throughput adds overhead and delay to the system without even taking into account internet routing delay.

Based on our research and related work, there is no intelligent, practical solution to replace the optimization approach and reduce the delay. Also, we found that no work has used DRL that has been demonstrated to be capable of tackling computational and combinatorial optimization issues, while considering current and predicted upcoming throughput.

The contributions of this paper can be summarized as follows:
\begin{itemize}
	\item We construct a regression model to address the link between the ROI size, Quality Factor (QF), and the total size of the video frame in bytes to compute the expected delay of the video frame.
	\item We also propose a DRL scheme that reduces the delay intelligently by adapting the ROI size and non-ROI quality based on the expected throughput.
	\item We integrate the DRL model with Web Real-Time Communication (WebRTC) framework \cite{vidgear} to stream and study the delay between two endpoints.
	
\end{itemize}

The remaining of the paper is arranged as follows. Section III provides an overview of our system's overall structure. Section IV discusses the proposed approach and each component of the system. In section V, the findings are displayed and discussed. Finally, conclusions are presented in Section VI.

\section{System Model}
\label{sec:model}
\begin{figure}[h]
	\centering
	\includegraphics[width=\linewidth]{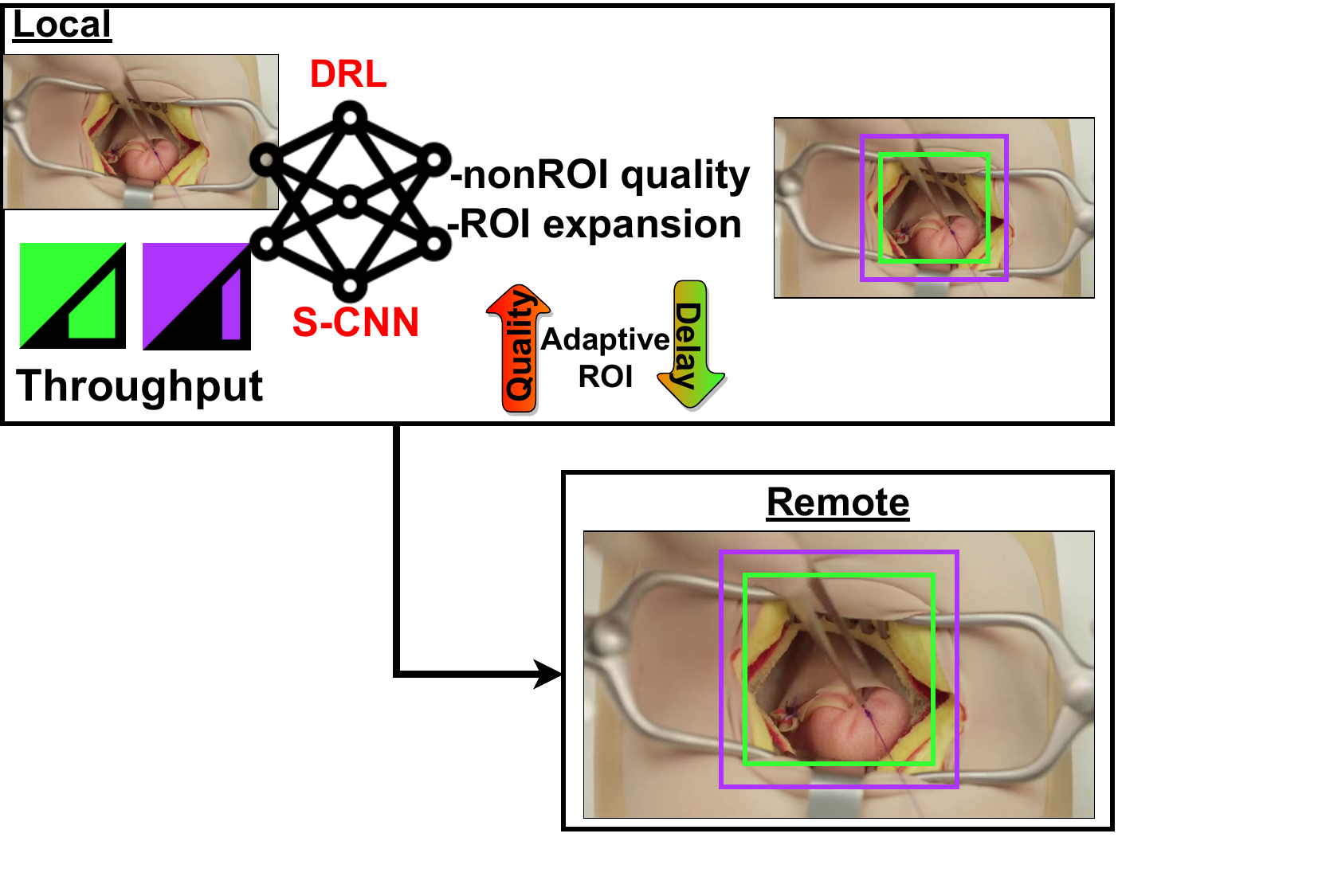}
	\caption{System Model}
	\label{fig_sys_model}
\end{figure}

\begin{figure}[H]
	\centering
	\includegraphics[width=\linewidth]{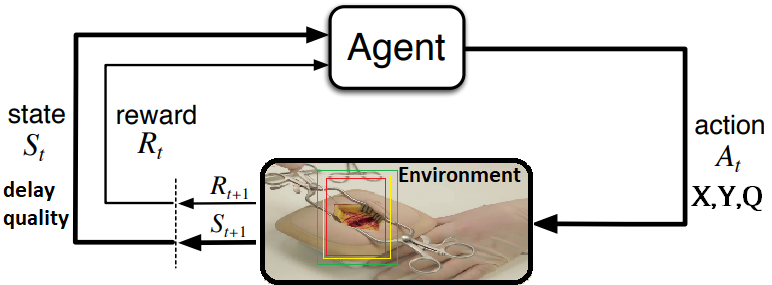}
	\caption{RL flowchart.}
	\label{fig:rl_flow_chart}
\end{figure}

Figure \ref{fig_sys_model} depicts the proposed system. The video frame is first sent to S-CNN to obtain the original ROI dimensions. We have used the S-CNN trained model from \cite{Somayya}. The ROI's width and height are calculated based on the detected coordination points. The architecture is made up of four convolutional layers, two max-pooling layers, and four fully connected layers. After getting the ROI dimensions, then it goes to the DRL model along with the current throughput. Depending on that throughput, the DRL model chooses the most effective ROI size and non-ROI quality factor that maximizes the quality score and minimizes the total delay. The essential premise of the RL is the Markov Decision Process (MDP), as shown in Figure \ref{fig:rl_flow_chart}, which comprises of four tuples \{Action, State, Transition, Reward\}. The agent chooses an action at time step ($t$), and the environment delivers a new state $S_{t+1}$ and a reward $R_{t+1}$. The agent's job is to figure out a policy, which is a mapping between the state and action, to maximize the expected value of its future rewards.

 \section{Proposed Approach}
\label{sec:problem}
The problem formulation for reducing the delay by modifying the ROI size and non-ROI quality is presented in this section. As highlighted in section \ref{sec:related}, the optimization problem described in \cite{Somayya} attempted to lower the latency while also improving the quality. However, this optimization problem did not take into consideration any dynamic changes in system throughput or latency. As a result, it was not adaptable in the time domain. In this section, we begin by obtaining throughput data, then DCT compression, followed by a description of the regression model, and finally, a discussion of the quality measurement tool.
\subsection{Calculating Throughput}
We collected throughput statistics using the Ookla speedtest framework \cite{ookla}, and used them as a dataset to train the DRL on throughput estimation. Throughput statistics were taken from our current location in Qatar to a remote server in Houston, Texas. The speed tests were carried out at various times of the day to ensure that the model covers the majority of the conceivable scenarios, with or without network congestion.
\subsection{DCT compression}
\begin{figure}
	\centering
	\begin{subfigure}{0.3\textwidth}
		\includegraphics[width=\linewidth]{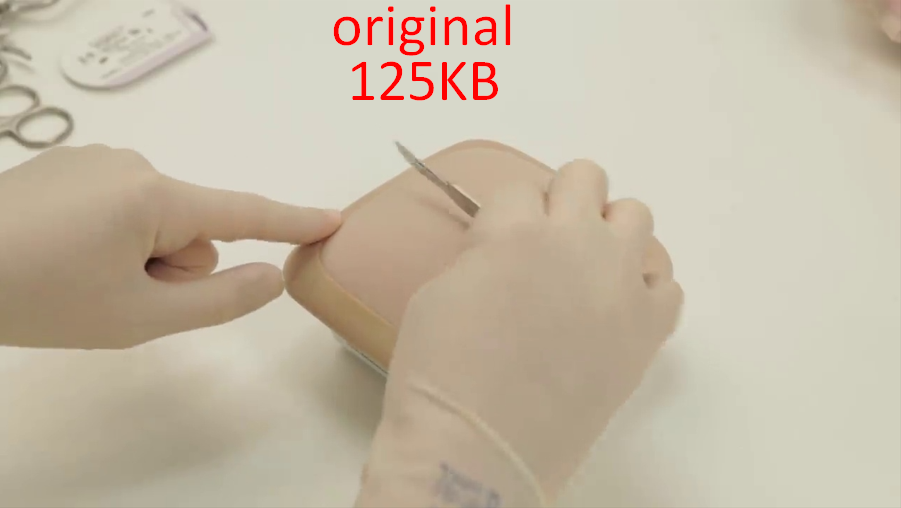}
		\caption{Original frame}
		\label{fig:original}
	\end{subfigure}
	\hfill
	\begin{subfigure}{0.3\textwidth}
		\includegraphics[width=\textwidth]{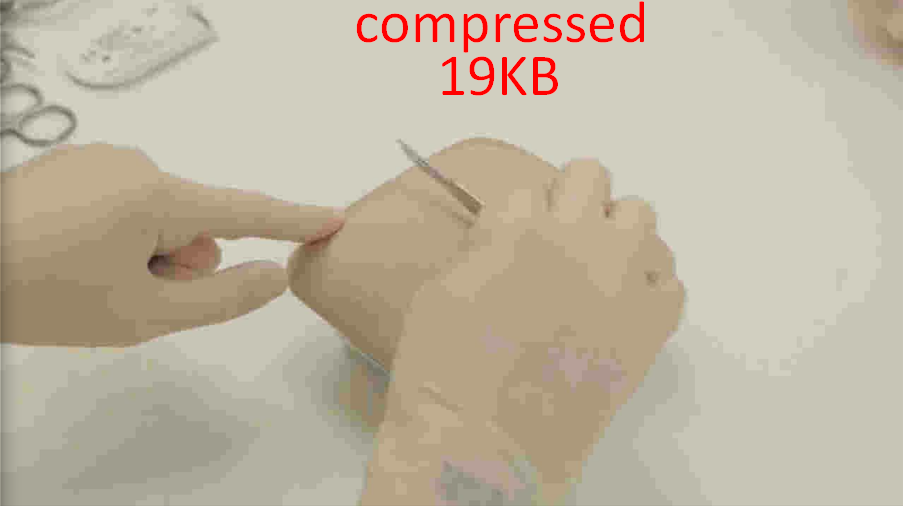}
		\caption{QF:10}
		\label{fig_qf10}
	\end{subfigure}
	\hfill
	\begin{subfigure}{0.3\textwidth}
		\includegraphics[width=\textwidth]{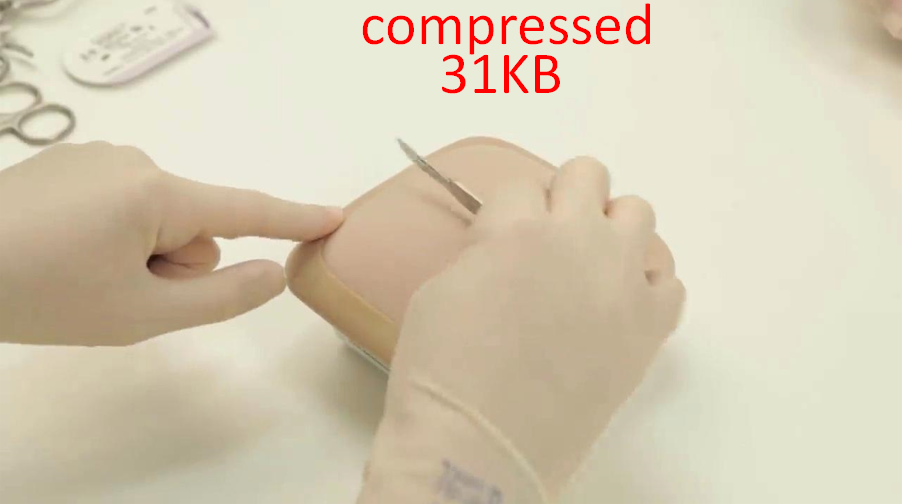}
		\caption{QF:100}
		\label{fig_qf100}
	\end{subfigure}

	\caption{Different QF comparison subfigures}
	\label{fig:QFCfigures}
\end{figure}

To lower the delay, frame size in bytes should be reduced, which necessitates frame compression. Doing so to the entire frame is not wise since specific telemedicine applications demand a clear and clean view to operate on. Therefore, compression usage will be directed at the non-ROI, which is used to identify the most critical part of the whole frame, and doing compression on the non-ROI will not affect its purpose \cite{Maglogiannis,Miaou}. DCT is a lossy compression that applies quantization to every 8x8 block of the image. Usually, the default JPEG quantization table is used to compress, but in our case, different quantization tables are needed depending on the throughput. Changing the quantization table is done using a function that produces the table with big coefficients when the QF is low and small coefficients when the QF is high. As shown in Figure \ref{fig_qf10}, the compression ratio of the DCT algorithm reaches up to 6.5 when QF is 10, and in Figure \ref{fig_qf100}, when QF is 100, the compression ratio reaches 4. After compression, the ROI portion is combined with the compressed part to obtain the whole frame. When the DRL model was deployed, the compression ratio for a four-minute video (246,060 KB) was 4.24.

\subsection{Regression Model}
\begin{figure}[H]
	\centering
	\includegraphics[width=\linewidth]{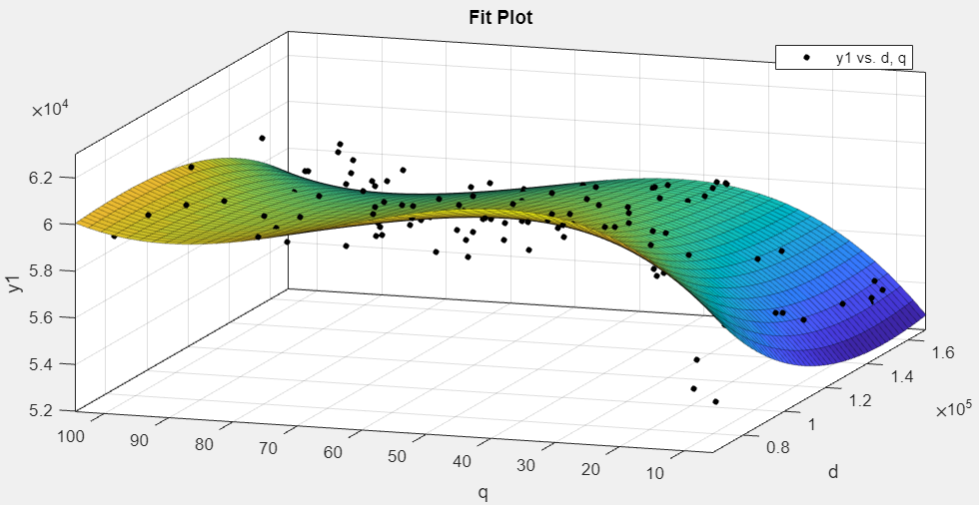}
	\caption{Regression model fit plot.}
	\label{fig:regressionfitplot}
\end{figure}

In order to calculate the delay, a regression model is needed to figure out the link between the total size of the frame and ROI size, and QF. To generate the model, we need to collect different random ROI, QF, and total frame sizes. Then the formula for 	  the regression model with cubic function which is polynomial of degree three should be as follows:
\begin{multline}
	S(d,q) = p_{00} + p_{10}\cdot x + p_{01}\cdot y + p_{20}\cdot x^{2} + p_{11}\cdot x\cdot y \\  + p_{02}\cdot y^{2} + p_{30}\cdot x^{3} + p_{21}\cdot x^{2}\cdot y + p_{12}\cdot x\cdot y^{2} + p_{03}\cdot y^{3}
	\label{eq:reg}
\end{multline}
where $S$ is the total size of the frame in bytes, $d$ is the area of the ROI $d=x\cdot y$, $x$ is the width of the ROI, $y$ is the height of the ROI and $q$ is the QF of the background.
As shown in figure \ref{fig:regressionfitplot}, using the MATLAB regression tool, with an R-Square of 0.822, the parameter values for the regression model are as follows:\\
	   $p_{00} =   6.256\times 10^4$,
       $p_{10} =     -0.2356$, 
       $p_{01} =       432.4$, 
       $p_{20} =   1.412\times 10^ {-6}$, 
       $p_{11} =    0.001398$,  
       $p_{02} =      -8.561$, 
       $p_{30} =  -2.637\times 10^{-12}$,  
       $p_{21} =   -8.87\times 10^ {-9}$,  
       $p_{12} =   6.147\times 10^ {-6}$,
       $p_{03} =     0.04034$. 

The delay is then calculated as shown in eq. (\ref{eq:delay}):
\begin{align}
	\gamma= S(d,q)/throughput
	\label{eq:delay}
\end{align}

\subsection{Quality Measurement}
Various tools can be used to assess the quality of the frame or image. Peak signal-to-noise ratio (PSNR) is the most well-known way to compare the original and compressed image. However, due to the change in the luminance and contrast in the image after compression, some low-quality frames have resulted in having a bigger PSNR than high-quality ones. This is an issue because the quality score will be used later in the DRL model since we want a higher score for better-quality frames. Therefore, structural
similarity index measure (SSIM) has shown to be a better metric for quality in all aspects, especially when there is a change in brightness \cite{Setiadi}.
SSIM is composed of three terms as shown in eq. (\ref{eq:ssim}). The comparison terms are luminance $l$, contrast $c$, and structure $s$ \cite{Hore}. These three terms will provide more accurate results when comparing the quality of original $f$ and compressed frames $g$.
\begin{align}
	\lambda(f,g) = l(f,g)\cdot c(f,g)\cdot s(f,g)
	\label{eq:ssim}
\end{align}

\subsection{Deep Reinforcement Learning}
DRL is the system's main component and is responsible for bringing intelligence to the system. It comprises different parts, which will be discussed in the following subsections.
\subsubsection{Environment Design}
The environment is simply a video frame that is being processed, and the observations/states of delay and quality are being calculated. The video frames used are the same ones in \cite{Somayya} and can be found in \cite{FundamentalVR,School}. As illustrated in Figure \ref{fig:rl_flow_chart}, each time step takes one frame, and then the model, depending on the state, will output a particular action from the action space that will be discussed in the following subsection.
\subsubsection{Action Spaces and States}
The action space is composed of three parameters representing the ROI dimensions $X$, $Y$, and the quality factor QF, A=\{$X$, $Y$, QF\}. Since the action space is continuous, we have used the min-max normalization so that the three parameters will be bounded between zero and one. This will help the model later to choose the action in the exploration phase and it improves training and prevents divergence \cite{elie}. In the beginning, the original ROI dimensions are obtained from the S-CNN ROI detection scheme. $X$ and $Y$ actions are the measures of how much the original width and length of the ROI should increase. It can reach up to its full frame size.
QF is the non-ROI quality, and it is similarly limited between zero and one, with one being the best quality.
Each state ($s$) in state set S has a delay, quality and throughput. $s$=\{$\gamma$, $\lambda$, $T$\} where ($\gamma$) is the delay and ($\lambda$) is the quality and ($T$) is the throughput. Delay is calculated by equation (\ref{eq:delay}) where the regression model $x$ input is the original ROI width + action ($X$), $y$ input is the original ROI height + action($Y$), $q$ input is the action (QF), and finally the current throughput ($T$) in the denominator.
\subsubsection{Reward Function}
The reward function restricts and guides the model's policy and actions to reach specific targets. Our reward function is defined to minimize the delay and maximize the quality, and anything opposite to that is considered as negative reward which is equivalent to a penalty.
As shown in Reward function in equation  (\ref{eq:rewardFunction}), we have selected throughput $103,076$ MB/s as a reference point, where we can tell the model that below this point we consider the throughput as low.

{\footnotesize
\begin{align}
    \mathbf{Reward}(\gamma, \lambda, T) :=
    \begin{cases}
        (\gamma^{-1} + \lambda^{-1}) & \text{if } T < 103,076 \\
        (\gamma + \lambda) & \text{if } T \geq 103,076
    \end{cases}
    \label{eq:rewardFunction}
\end{align}}

\subsubsection{Deep Reinforcement Learning Algorithm}
The agent's objective is to figure out the policy that maximizes the total discounted reward. Because original RL algorithms cannot handle continuous actions, we can use either Deep Deterministic Policy Gradient (DDPG) or Soft Actor Critic (SAC). We have selected SAC algorithm since it was stable for our environment and demonstrated better results. Another reason is that authors in \cite{Kei} concluded that DDPG performs well with some non-complex continuous action spaces but lacks a quicker convergence rate.
SAC is an algorithm that optimizes a stochastic policy in an off-policy manner, bridging the gap among stochastic policy optimization and DDPG-style methods \cite{Tuomas}.
SAC is also known that it trains and explores a stochastic policy with entropy regularization, and explores in an on-policy way. It is also follows actor-critic architecture with separate policy and value function networks.
Generally, SAC algorithm as shown in algorithm \ref{alg:soft_actor_critic}, aims to learn the policy $\policy_\pparams(a_t|s_t)$, Soft Q-value $\Q_\params(s_t, a_t)$ and soft state value  $\V_\vparams(s_t)$.
The Soft Q-value function (\ref{eq:sac3}) is obtained from Bellman equation (\ref{eq:sac1}) and soft state value (\ref{eq:sac2}).
\begin{align}
Q(s_t, a_t) &= r(s_t, a_t) + \gamma \mathbb{E}_{s_{t+1} \sim \rho_{\pi}(s)} [V(s_{t+1})]
\label{eq:sac1}
\end{align}
\begin{align}
V(s_t) &= \mathbb{E}_{a_t \sim \pi} [Q(s_t, a_t) - \alpha \log \pi(a_t \vert s_t)]
	\label{eq:sac2}
\end{align}
\begin{multline}
Q(s_t, a_t) = r(s_t, a_t) + \gamma \mathbb{E}_{(s_{t+1}, a_{t+1}) \sim \rho_{\pi}} [Q(s_{t+1}, a_{t+1}) \\ - \alpha \log \pi(a_{t+1} \vert s_{t+1})]
	\label{eq:sac3}
\end{multline}
The parameters of Q-function and the policy networks are $\vparams,\ \params$, $\pparams$, and $\bar\psi$;  $\epsilon_t$ is an input noise vector.
The gradient in equation (\ref{eq:v_gradient}) is used to train the soft value function to minimize the squared residual error.
\setlength{\abovedisplayskip}{1pt}
\setlength{\belowdisplayskip}{1pt}
\begin{align}
\resizebox{\columnwidth}{!}{$
\hat \nabla_\vparams J_V(\vparams) = \nabla_\vparams \V_\vparams(s_t) \left(\V_\vparams(s_t) - Q_\params(s_t, a_t) + \log \policy_\pparams(a_t|s_t)\right),$}
\label{eq:v_gradient}
\end{align}
Also by using stochastic gradients as shown in equation (\ref{eq:q_gradient}), soft Q-function parameters can be trained to minimize and optimize the soft Bellman residual.

\begin{align}
\resizebox{\columnwidth}{!}{$
\hat \nabla_\params J_Q(\params) =  \nabla_\params \Q_\params(a_t, s_t) \left(\Q_\params(s_t, a_t) - \reward(s_t, a_t) - \discount \V_\vtargetparams(\stp)\right)$}.
\label{eq:q_gradient}
\end{align}

\begin{algorithm}[h]
\caption{Soft Actor-Critic}
\label{alg:soft_actor_critic}
\begin{algorithmic}
\State \mbox{Initialize parameter vectors $\vparams$, $\vtargetparams$, $\params$, $\pparams$.}  \\
\textbf{for} each iteration \textbf{do}  \\
	\quad\textbf{for} each environment step \textbf{do}
	    \State \quad\quad $a_t \sim \policy_\pparams(a_t|s_t)$
	    \State \quad\quad $\stp \sim \pdyn(\stp| s_t, a_t)$
	    \State \quad\quad $\mathcal{D} \leftarrow \mathcal{D} \cup \left\{(s_t, a_t, \reward(s_t,a_t),\stp)\right\}$ \\\quad\textbf{end for}\\
	\quad\textbf{for} each gradient step \textbf{do}
	    \item \quad\quad $\vparams \leftarrow \vparams - \lambda_V \hat \nabla_\vparams J_\V(\vparams)$
    	\State \quad\quad $\params_i \leftarrow \params_i - \lambda_Q \hat \nabla_{\params_i} J_\Q(\params_i)$ for $i\in\{1, 2\}$
    	\State \quad\quad $\pparams \leftarrow \pparams - \lambda_\policy \hat \nabla_\pparams J_\policy(\pparams)$
    	\State \quad\quad $\vtargetparams\leftarrow \tau \vparams + (1-\tau)\vtargetparams$
\\\quad\textbf{end for}
\\ \textbf{end for}
\end{algorithmic}
\end{algorithm}

\section{Results}
\label{sec:results}
We explain the results of training the DRL model and the new changes in latency and quality after implementing the model, and compare the performance of the proposed solution with the state-of-the-art.
\subsection{DRL Convergence}
\begin{figure}[H]
	\centering
	\includegraphics[width=\linewidth]{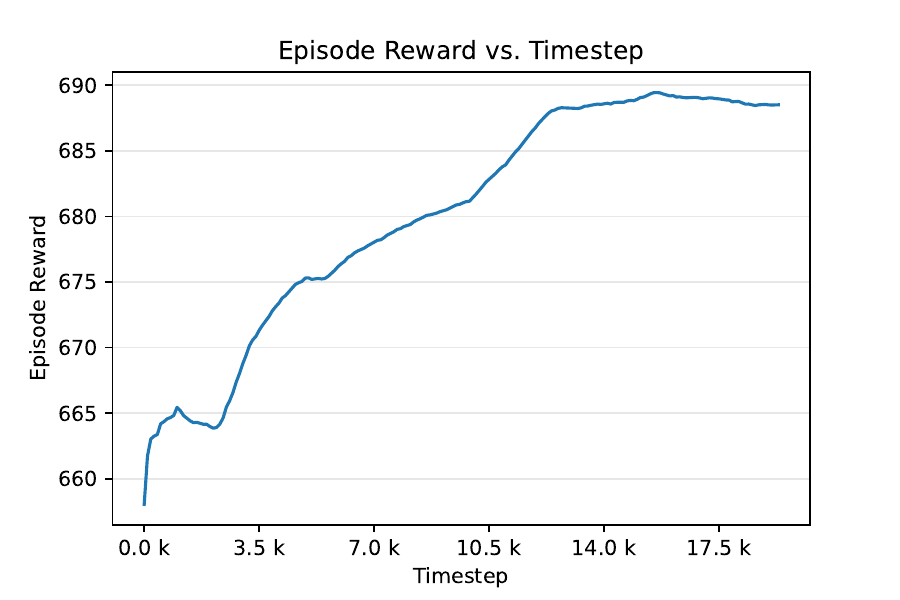}
	\caption{DRL episode reward convergence during the training.}
	\label{fig_;learning_curve}
\end{figure}

As illustrated in Figure \ref{fig_;learning_curve}, the average episode reward converged to roughly 688 after 20,000 timesteps of training at a 0.002 learning rate.

\subsection{Delay Comparison Results}
\begin{figure}[h]
	\centering
	\includegraphics[width=\linewidth]{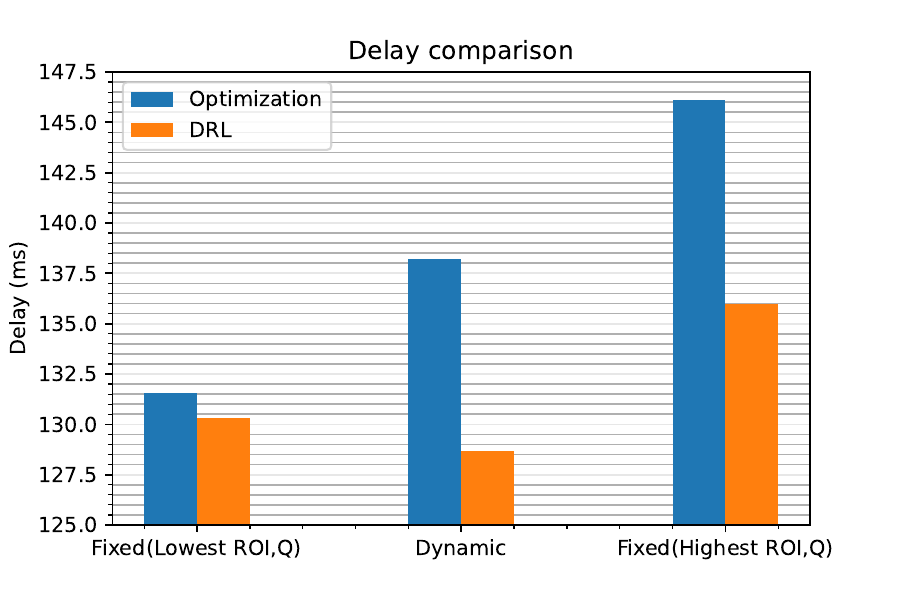}
	\caption{Delay comparison between different options}
	\label{fig_delay}
\end{figure}

\begin{figure}[h]
	\centering
	\includegraphics[width=\linewidth]{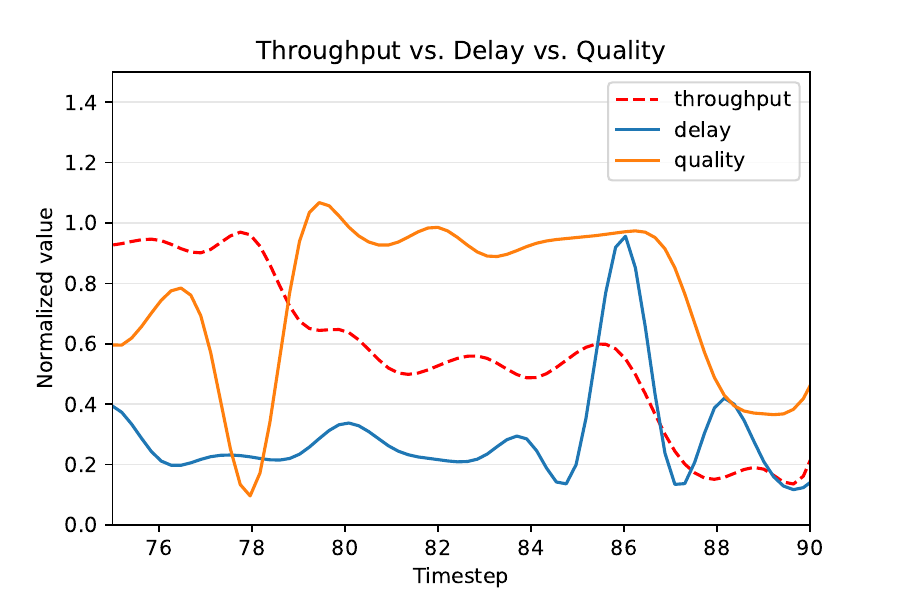}
	\caption{Throughput vs. Delay vs. Quality}
	\label{fig_dtq}
\end{figure}

The findings of the delay demonstrate that DRL has promise and can reduce the delay and enhance the interactivity of the system. Using the DRL in telemedicine applications can open up a slew of new doors in the medical world. Figure \ref{fig_delay} depicts that DRL surpasses both the fixed lowest ROI, quality and the fixed highest ROI, quality. We also compared our work to that in \cite{Somayya}, and as shown in the figure, the delays were reduced by nearly 13\%. These results did not account for the streaming overhead, which is expected to add additional delay. Figure \ref{fig_dtq} depicts dynamic changes in delay and quality. When throughput decreases, delay and quality gradually decrease to react to the change in throughput and remain in a trade-off situation. A min-max normalization has been used to make the comparison clear and readable. The min and max values for throughput, delay, and quality are as follows:  
\begin{align}
throughput:\{1.7912,9.5001\}\\
delay:\{0.0791,0.2541\}\\ 
quality:\{0.6144,0.9839\} 
\end{align}

\subsection{Quality Comparison Results}
\begin{figure}[h]
	\centering
	\includegraphics[width=\linewidth]{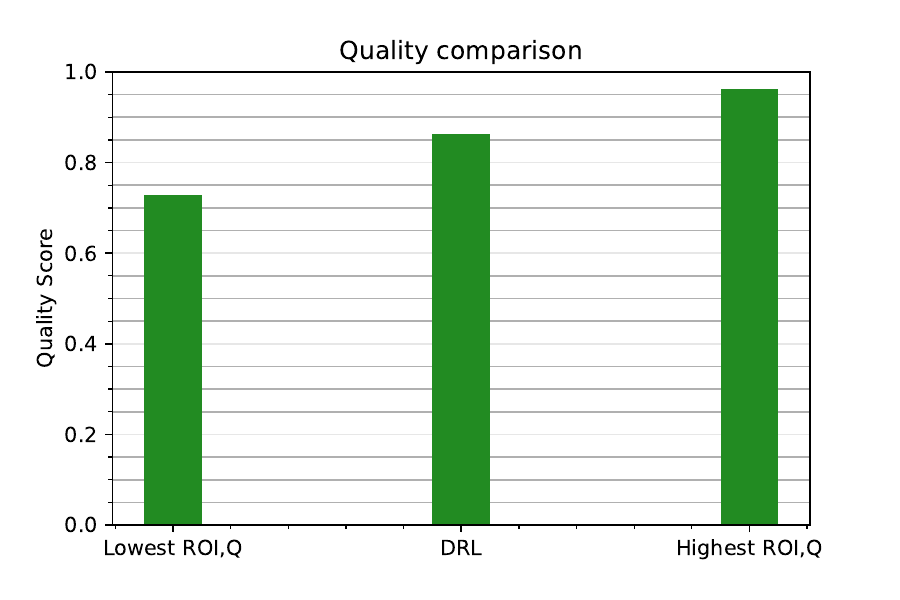}
	\caption{Quality comparison between different options}
	\label{fig_quality}
\end{figure}

Figure \ref{fig_quality} shows quality findings obtained using the SSIM metric. DRL received a score of 0.86 out of 1, putting it in between the low-quality non-ROI and the high-quality non-ROI options.

\subsection{WebRTC Results}
\begin{figure}[h]
	\centering
	\begin{subfigure}{0.3\textwidth}
		\includegraphics[width=\linewidth]{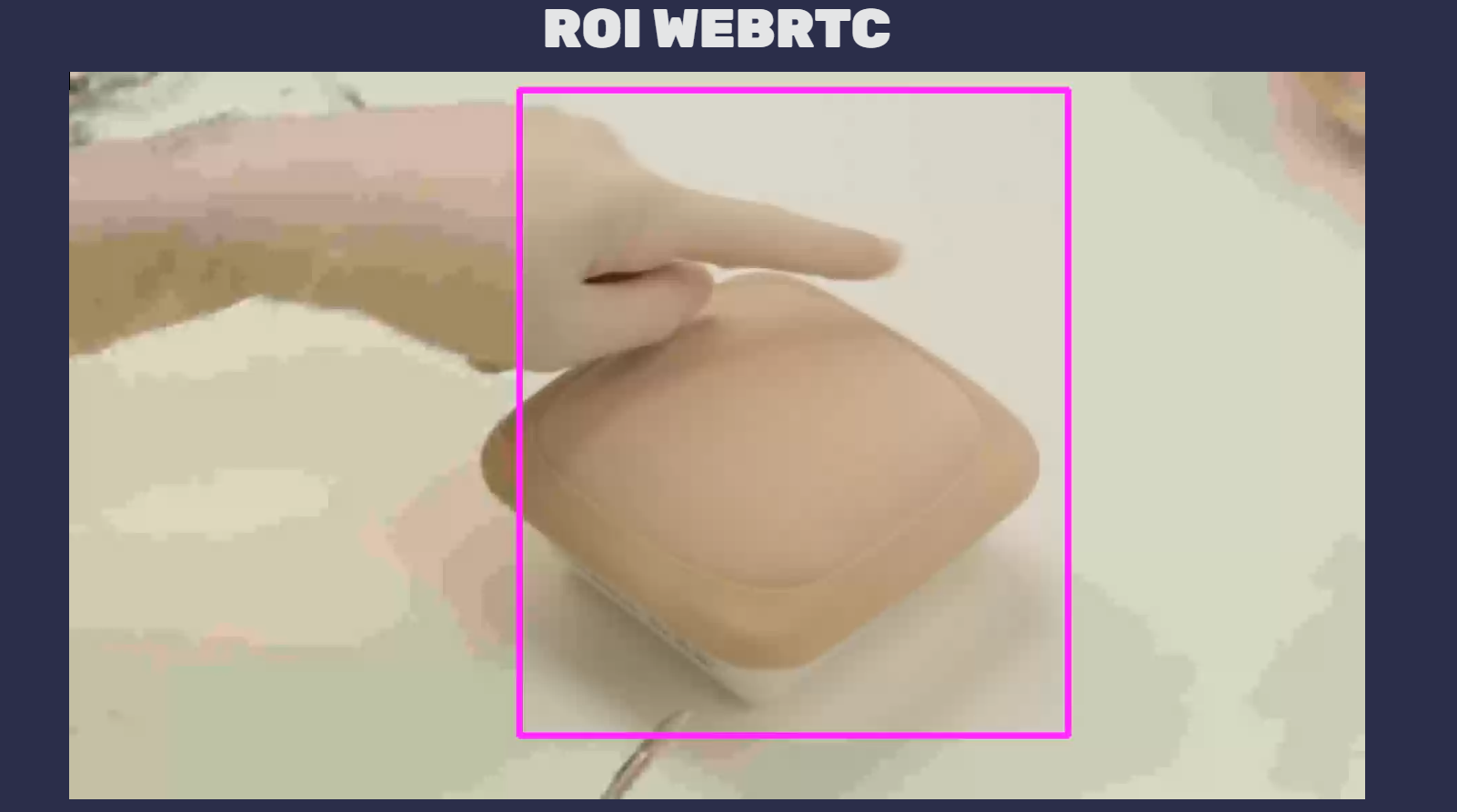}
		\caption{throughput is low 1.7912 Mb/s}
		\label{fig:low_webrtc}
	\end{subfigure}
	\hfill
	\begin{subfigure}{0.3\textwidth}
		\includegraphics[width=\textwidth]{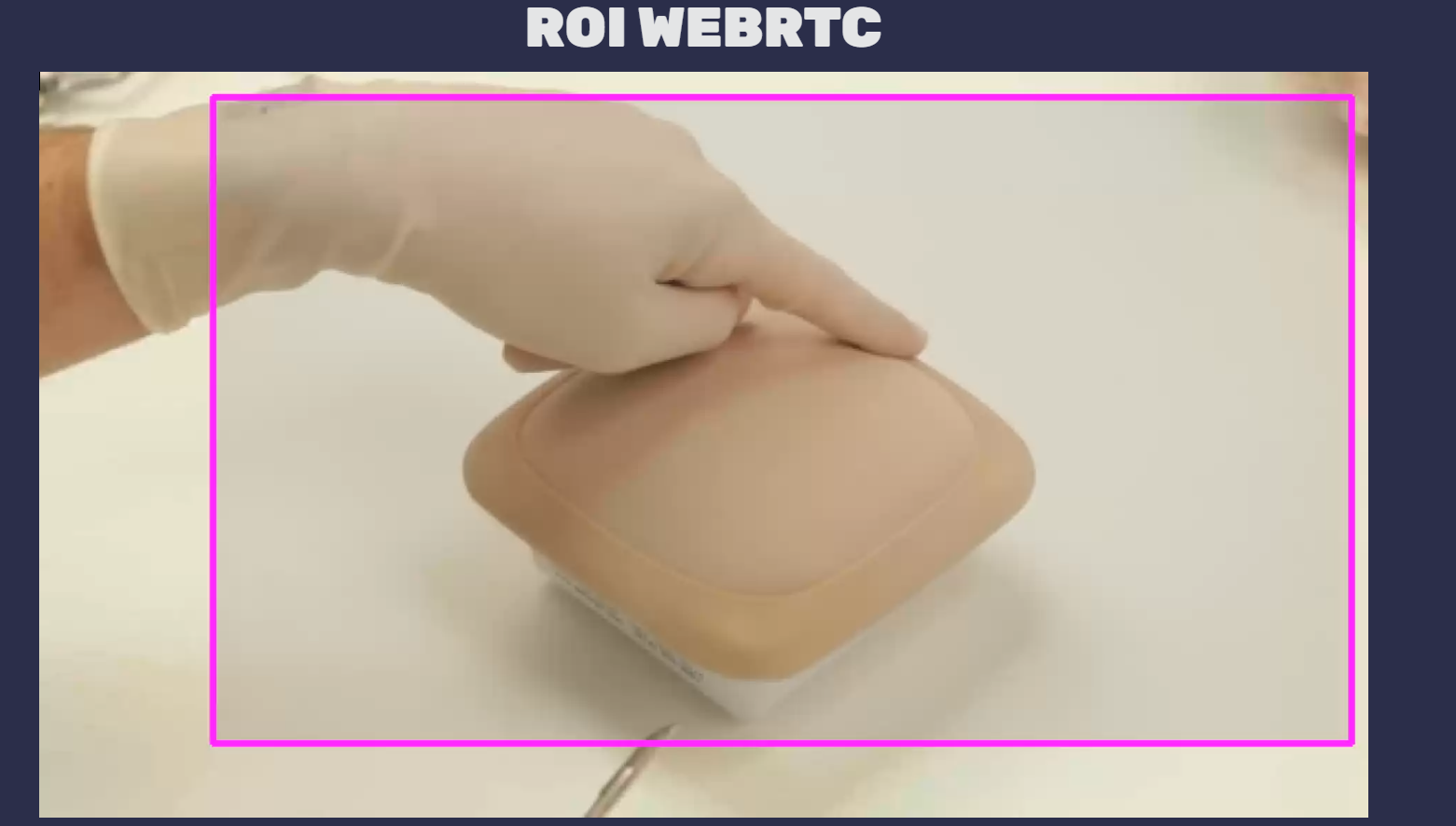}
		\caption{throughput is high 9.50008 Mb/s}
		\label{fig:high_webrtc}
	\end{subfigure}
	
	\caption{Comparison WebRTC streaming implementation with different throughput values.}
	\label{fig:webrtcsubfigures}
\end{figure}

Applying the model to a real-world example allows comparing it to existing solutions to the current research problem. We have implemented surgical video streaming example using WebRTC technology, and we integrated the DRL model with it.
Figures \ref{fig:low_webrtc}, \ref{fig:high_webrtc} provide examples of implementation when the throughput is low and high, respectively.

\begin{figure}[H]
	\centering
	\includegraphics[width=\linewidth]{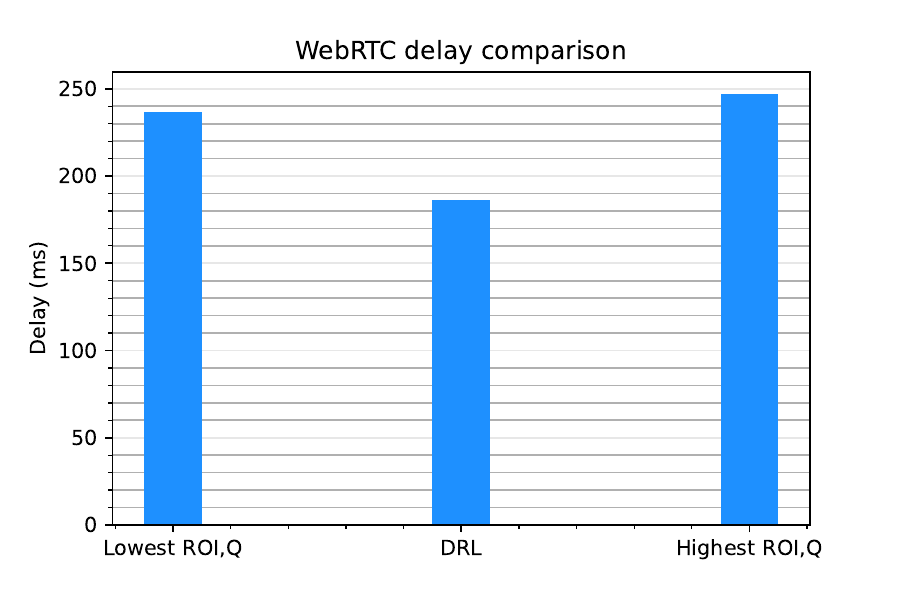}
	\caption{WebRTC Delay comparison }
	\label{fig_webrtc-quality}
\end{figure}

As seen in Figure \ref{fig_webrtc-quality}, the average delay for the full video in the WebRTC implementation showed that there is 33\% delay enhancement compared to the highest ROI and quality. After comparing the theoretical and practical results that reduced the delay while maintaining the quality, we can clearly say that DRL is a viable approach to reduce the delay. It is foreseen that it will become a common approach for reducing delay in telemedicine applications.

\section{Conclusion}
\label{sec:conc}
In this paper, we proposed a DRL model that adapts the ROI size and non-ROI quality based on the projected throughput. The DRL model was evaluated using delay and quality tools. The findings demonstrate that the model helps reduce the delay by 13\%, while maintaining the quality of the non-ROI. 
Lowering the latency and attaining real-time telemedicine applications will continue to be a prevalent issue and study field. Future work may be required to reduce the compression processing time by adding prediction models to the upcoming frames, also using intelligent routing to the video packets would help reducing the delay more.

\section*{Acknowledgment}
This work was supported by NPRP award (NPRP12S-0119-190006) from the Qatar National Research Fund (a member of The Qatar Foundation). The findings achieved herein are solely
the responsibility of the authors.

\bibliographystyle{IEEEtran}
\bibliography{biblio}

\end{document}